# SAR Image Change Detection via Spatial Metric Learning With an Improved Mahalanobis Distance

Rongfang Wang, *Member, IEEE*, Jia-Wei Chen, *Member, IEEE*, Yule Wang, Licheng Jiao, *Fellow, IEEE*, and Mi Wang

*Abstract*—The log-ratio (LR) operator has been widely employed to generate the difference image for synthetic aperture radar (SAR) image change detection. However, the difference image generated by this pixelwise operator can be subject to SAR images speckle and unavoidable registration errors between bitemporal SAR images. In this letter, we proposed a spatial metric learning method to obtain a difference image that is more robust to the speckle by learning a metric from a set of constraint pairs. In the proposed method, the spatial context is considered in constructing constraint pairs, each of which consists of patches in the same location of bitemporal SAR images. Then, a semidefinite positive metric matrix M can be obtained by the optimization with the max-margin criterion. Finally, we verify our proposed method on four challenging data sets of bitemporal SAR images. Experimental results demonstrate that the difference map obtained by our proposed method outperforms than other state-of-the-art methods.

## I. INTRODUCTION

CHANGE detection is the process of identifying differences in the state of an object or phenomenon by observing it at various time points [1], which has been widely applied in the evaluation of the attack effects in combatant areas, the survey of geographical information in a certain region, and monitor atmospheric effects. In past decades, remote sensing images have been widely used for change detection.

A synthetic aperture radar (SAR) is a sensor that is not limited by seasons or the weather in its imaging operations.

Manuscript received January 3, 2019; revised March 8, 2019; accepted May 3, 2019. This work was supported in part by the National Natural Science Foundation of China under Grant 61701361 and Grant 61806154, in part by the Open Fund of State Laboratory of Information Engineering in Surveying, Mapping and Remote Sensing, Wuhan University, under Grant 17E02, in part by the Natural Science Basic Research Plan in Shaanxi Province of China under Grant 2018JM6083, and in part by the Fundamental Research Funds for the Central University under Grant JB181701. *(Corresponding author: Jia-Wei Chen.)*

R. Wang, J.-W. Chen, Y. Wang, and L. Jiao are with the Key Laboratory of Intelligent Perception and Image Understanding of Ministry of Education, School of Artificial Intelligence, Xidian University, Xi'an 710071, China (e-mail: rfwang@xidian.edu.cn; jawaechan@gmail.com).

M. Wang is with the State Key Laboratory of Information Engineering in Surveying, Mapping and Remote Sensing, Wuhan University, Wuhan 430079, China.

Color versions of one or more of the figures in this letter are available online at http://ieeexplore.ieee.org.

Digital Object Identifier 10.1109/LGRS.2019.2915251

It transmits microwaves to the surface of the Earth and receives the echo data reflected from the ground, which are used to generate a SAR image. Therefore, SAR image data contain a great deal of useful information about the terrain of the earth and they have been popularly employed for change detection due to their many advantages. Recently, Cui *et al.* [2] proposed a benchmark of similarity measurement for SAR image change detection.

At present, most of the SAR image change detection methods are developed based on the framework proposed in [3] and [4], in which the changed regions are detected from a difference image. As speckle is modeled as multiplicative noise, Bovolo and Bruzzone [5] proposed a difference image based on a log-ratio (LR) operator. However, this pixelwise operator is subject to SAR image speckle and unavoidable registration errors between bitemporal SAR images, which will reduce the accuracy of change detection. This issue can be tackled in two ways. First, robust features are extracted from the difference image to reduce the influence on the accuracy of change detection. Zhang *et al.* [6] proposed a graph-cut method to extract the change region based on the LR difference image. Gong *et al.* [7] proposed a fuzzy clustering method based on the Markov random field (MRF). Wang *et al.* [8] discussed the effects of despeckling and structure features on SAR image change detection. Recently, the methods based on deep learning have been developed especially for change detection [9]–[12]. The common goal of these methods is to learn robust discriminative features on changed regions of the difference image.

Alternatively, prior works are employed to generate a robust difference image. Gong *et al.* [13] proposed a neighborhood-based LR operator, where the spatial context information is considered with the difference image to reduce the influence of speckle and registration errors. In fact, the similarity measurement is a fundamental issue for SAR image change detection. Cui *et al.* [2] summarized a variety of similarity measurements for change detection in multiple temporal SAR images, where different similarities are developed for different types of change. Recently, learning-based methods were proposed to represent the difference between multiple temporal SAR images. Liu *et al.* [14] proposed a method based on deep neural network (DNN) to learn a representation of the difference between temporal SAR images, where restricted Boltzmann machines (RBMs) are stacked to form a deep hierarchical neural network to learn the difference image and recognize changed and unchanged regions. With the similar idea, Zhang et al. [15] proposed a DNN to learn discriminative difference representations (DRs) for different types of changes. Being different from the previous network, the latter network is trained without any ground truth. Instead, it is trained using pseudolabels generated via clustering to obtain robust features, and the robust features are subsequently applied to update the pseudolabel via clustering.

Most of the aforementioned methods focus on learning or extracting the features of the changed and unchanged regions and using these features to recognize changed regions. However, in most cases, the feasible feature-based description of the changed region might be difficult to obtain due to semantic gaps. In recent decades, metric learning has been developed to learn distance metrics from data and it has been widely employed in image classification and data analysis [16]. The goal of metric learning is to obtain a measurement of the (dis)similarity of the objects to be classified. In some cases, the (dis)similarities will simplify the issue. Metric learning primarily aims to learn a positive semidefinite (PSD) matrix $\mathbf{M}$ from two pairwise constraint sets: a must-link constraint set and a can-not-link constraint set. Then, the trained matrix $\mathbf{M}$ is applied to measure the similarity of a pair of data samples with the form of a Mahalanobis metric [17]. Do et al. 18 proposed a metric learning with the structured support vector machine (SVM). The idea of metric learning can be naturally applied to SAR image change detection to learn a distance metric measuring the difference between a pair of temporal SAR images.

Inspired by this, in this letter, we propose a spatial metric learning method to learn a difference image from bitemporal SAR images, which is robust to the effects of the speckle and the unavoidable registration errors. To achieve this, in the proposed method, we learn a PSD metric matrix that is learned from a set of constraint pairs with spatial context information to reduce the effect of registration errors. More specifically, each constraint pair is constrained from the patches around the same location of bitemporal SAR images to measure their difference. Then, the metric matrix $\mathbf{M}$ can be optimized with a maximal margin criterion. Finally, the proposed method is verified on several sets of multiple temporal SAR images. The experimental results show that a more robust difference image is obtained by our proposed method. Furthermore, several methods are compared with our proposed method, and the results show that it outperforms the other methods.

The rest of this letter is organized as follows. The proposed algorithm will be introduced in detail in Section II. Experimental results will be presented in Section III. Finally, a short conclusion will be made in Section IV.

## II. Proposed Method

In this section, a brief review of metric learning is given and then a spatial metric learning method will be introduced for temporal SAR image change detection.

### A. Metric Learning

As mentioned above, metric learning aims to learn a measurement of (dis)similarity among the objects to be classified. This means that given a set of samples $(\mathbf{x}_1, \mathbf{x}_2, \ldots, \mathbf{x}_N)$, $\mathbf{x}_i \in \mathbb{R}^d$, the goal is to learn a PSD matrix $\mathbf{M}$ that is used to parameterize the Mahalanobis distance [17]

$$d_\mathbf{M}(\mathbf{x}_i, \mathbf{x}_j) = (\mathbf{x}_i - \mathbf{x}_j)^T \mathbf{M}(\mathbf{x}_i - \mathbf{x}_j). \tag{1}$$

With this parameterized Mahalanobis distance, we can measure the similarity or the distance between two data samples $\mathbf{x}_i$ and $\mathbf{x}_j$ with multivariate Gaussian distributions.

Usually, given two constraint sets: a must-link constraint set $\mathbf{L}$ and a cannot-link constraint set $\mathbf{C}$, which can be defined as follows:

$$\mathcal{L} : \{(\mathbf{x}_i, \mathbf{x}_j) | \mathbf{x}_i \text{ is similar to } \mathbf{x}_j\}$$
$$\mathcal{C} : \{(\mathbf{x}_i, \mathbf{x}_j) | \mathbf{x}_i \text{ is dissimilar to } \mathbf{x}_j\}. \tag{2}$$

The goal of metric learning is to train a mapping function $\phi(\mathbf{x}_i, \mathbf{x}_j)$ that maps a constraint pair to a variable, which indicates whether the pair of samples is similar. More specifically, mapping function $\phi(\mathbf{x}_i, \mathbf{x}_j)$ is usually defined by (1), and the training of the mapping function $\phi(\mathbf{x}_i, \mathbf{x}_j)$ can be essentially converted to the estimation of the PSD matrix $\mathbf{M}$ using the constraint pairs.

Many metric learning methods have been developed. Xiong et al. [19], [20] proposed two metric learning methods based on the random forest and max-margin criterion, respectively. These two methods convert metric learning into a binary classification task and construct the constraint pairs using the samples belonging to the same class.

### B. Spatial Metric Learning for Change Detection

Most conventional change detection methods were developed based on the difference map generated by the LR operator. Instead, in this section, we propose spatial metric learning for temporal SAR image change detection. In contrast to conventional methods, the difference information can be learned from bitemporal SAR images.

Given bitemporal SAR images $\mathbf{I}_1$ and $\mathbf{I}_2$, they have been approximately registered. For each site, patches $\mathbf{p}_i$ and $\mathbf{q}_i$ from these bitemporal SAR images are extracted as a pair of constraints, which will be used to learn a metric for measuring the difference between these bitemporal SAR images. However, it exists usually as unavoidable registration errors between the aligned bitemporal SAR images, and the constraint pairs derived by the patches in the same location are noisy. In addition, the number of pair constraints derived in this manner is quite limited due to a small number of available labeled pixels. The metric trained by these noisy and insufficient constraint pairs will inevitably perform poorly, and moreover, the metric can even perform worse than the straightforward Euclidean distance metric [21].

To solve this issue, a spatial context prior is considered in the construction of constraint pairs. To achieve this, for each location, a patch $\mathbf{p}_i$ is extracted from SAR image $\mathbf{I}_1$. Then, in the neighborhood of the current site, a set of patches $\mathbf{Q}_i$ is extracted from SAR image $\mathbf{I}_2$, from which a patch $\mathbf{q}_i$, with the different landcovers from the patch $\mathbf{p}_i$ will be added into a positive constraint pair. Otherwise, patch $\mathbf{q}_i$ will be added into a negative constraint pair with patch $\mathbf{p}_i$. The entire pipeline is illustrated in Fig. 1.

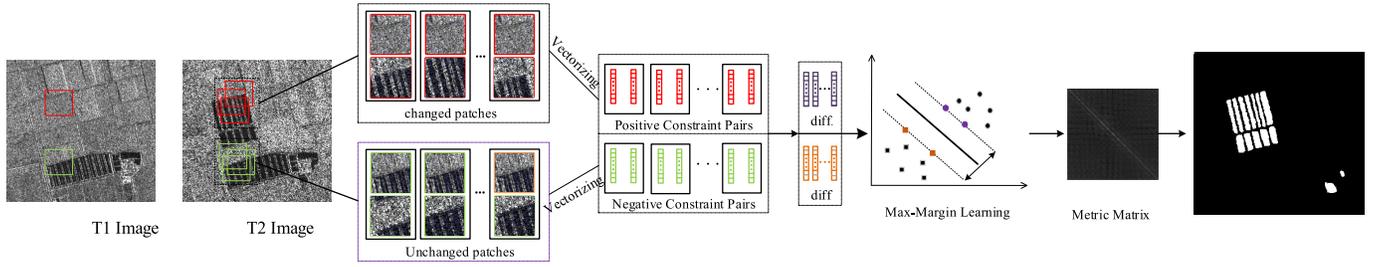

Fig. 1. Spatial metric learning framework for SAR image change detection.

So far, a constraint set has been constructed to learn a metric for bitemporal SAR images change detection, and it includes the positive constraint pairs $\mathcal{L}$ for the changed class and the negative constraint pairs $\mathcal{C}$ for unchanged class. In the constraint set, for each pair of constraints $(\mathbf{x}_{k1}, \mathbf{x}_{k2})$, a more general form of the Mahalanobis distance can be employed to measure their difference

$$d_{\mathbf{M}}(\mathbf{x}_{k1}, \mathbf{x}_{k2}) = \mathbf{v}_k^T \mathbf{M} \mathbf{v}_k \quad (3)$$

where $\mathbf{v}_k = \text{diff}(\mathbf{x}_{k1}, \mathbf{x}_{k2})$ is a difference vector and the function $\text{diff}(\cdot, \cdot)$ indicates the difference of $\mathbf{x}_{k1}$ and $\mathbf{x}_{k2}$. When the function $\text{diff}(\cdot, \cdot)$ is defined as

$$\text{diff}(\mathbf{x}_{k1}, \mathbf{x}_{k2}) = (\mathbf{x}_{k1} - \mathbf{x}_{k2}) \quad (4)$$

(3) is equivalent to (1). In this letter, due to the statistics of the speckle noise, we use the LR as the difference of each constraint pair, that is,

$$\text{diff}(\mathbf{x}_{k1}, \mathbf{x}_{k2}) = \log(\mathbf{x}_{k1} \odot \mathbf{x}_{k2}) \quad (5)$$

where the symbol $\odot$ denotes the elementwise division.

As the goal of the metric learning is to learn a PSD matrix $\mathbf{M}$, Xiong et al. [20] proposed a method to learn the PSD matrix based on the max-margin criterion. In fact, the metric learning model based on the max-margin can be expressed as follows:

$$\min_{\mathbf{M}, b, \epsilon} \frac{1}{2}\|\mathbf{M}\|_F^2 + \frac{C}{N}\sum_{k=1}^{N}\epsilon_k$$
$$\text{s.t. } y_k \cdot (d_{\mathbf{M}}(\mathbf{x}_{k1}, \mathbf{x}_{k2}) + b) \geq 1 - \epsilon_k$$
$$\epsilon_k > 0, \quad k = 1, 2, \ldots, N \quad (6)$$

where $\epsilon$ is the vector of relax variables $(\epsilon_1, \epsilon_2, \ldots, \epsilon_N)$, and $b$ is a bias. $y_k$ is a variable that indicates whether the $k$th constraint pair $(\mathbf{x}_{k1}, \mathbf{x}_{k2})$ is positive or negative one. $N$ is the number of constraint pairs for training and $C$ is a regularize parameter.

The model in (6) is a nonconvex optimal problem. To solve this, the original problem can be rewritten using vectorization as follows:

$$\min_{\mathbf{w}, \epsilon} \frac{1}{2}\|\mathbf{w}\|_2^2 + \frac{C}{N}\sum_{k=1}^{N}\epsilon_k$$
$$\text{s.t. } y_k \cdot \mathbf{w}^T \mathbf{u}_k \geq 1 - \epsilon_k$$
$$\epsilon_k > 0, \quad k = 1, 2, \ldots, N \quad (7)$$

where $\mathbf{w} = (b, \text{vec}(\mathbf{M})^T)^T$, $\mathbf{u}_k = (1, \text{vec}(\mathbf{v}_k \mathbf{v}_k^T)^T)^T$, and $\text{vec}(\cdot)$ is the operator that vectorizes a matrix into a column vector with columnwise arrangement. Then, the problem in (7) can be efficiently solved by using binary SVM classifier [22] with only one slack variable. To achieve this, the model in (7) can be converted to

$$\min_{\mathbf{w}, \epsilon} \frac{1}{2}\|\mathbf{w}\|_2^2 + C \cdot \epsilon$$
$$\text{s.t. } \forall \mathbf{c} \in \{0, 1\}^N, \frac{1}{N}\mathbf{w}^T \sum_{k=1}^{N} c_k y_k \mathbf{u}_k \geq \frac{1}{N}\sum_{k=1}^{N} c_k - \epsilon$$
$$\epsilon > 0 \quad (8)$$

where $\mathbf{c} = (c_1, c_2, \ldots, c_N)$ is an auxiliary vector in which each component $c_k \in \{0, 1\}$ is a variable indicating whether the $k$th constraint in (7) is selected in a working set. Then, with vector $\mathbf{c}$, the working set is a subset of constraints in (7). In addition, $\epsilon$ is the slack variable shared by all the constraints which measures training loss. It has been proven [22] that the models in (7) and (8) have the same solutions when $\epsilon = (1/N)\sum_{k=1}^{N}\epsilon_k$ is satisfied. Finally, the model in (8) can be efficiently solved by the cutting-plane algorithm [23].

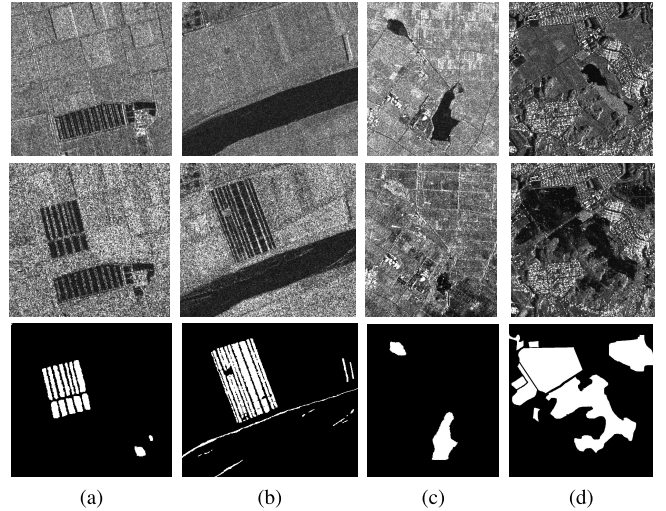

Fig. 2. Four sets of bitemporal SAR images. The first row is the images acquired in the first time and the second row is the images acquired in the second time. The last row is the ground truth. (a) Yellow River scene A. (b) Yellow River scene B. (c) Sendai earthquake scene A. (d) Sendai earthquake scene B.

## III. EXPERIMENTAL RESULTS AND ANALYSIS

In this letter, the proposed method is verified on four sets of bitemporal SAR images. Two scenes (YR-A and YR-B) are from bitemporal Yellow River SAR images [10] acquired by the Radarsat-2 satellite in 2008 and 2009, respectively. Their image sizes are $306 \times 291$ and $400 \times 350$, respectively. Other two are parts of TerraSAR-X images acquired prior to (on October 20, 2010) and after (on May 6, 2011) the Sendai earthquake in Japan [2]. Their image sizes are $590 \times 687$ and $689 \times 734$, respectively. These four data sets are shown in Fig. 2). These four data sets are quite challenging, such as the linear-shape changed regions in YR-B data set and complex scene in both Sendai-A and Sendai-B data sets.

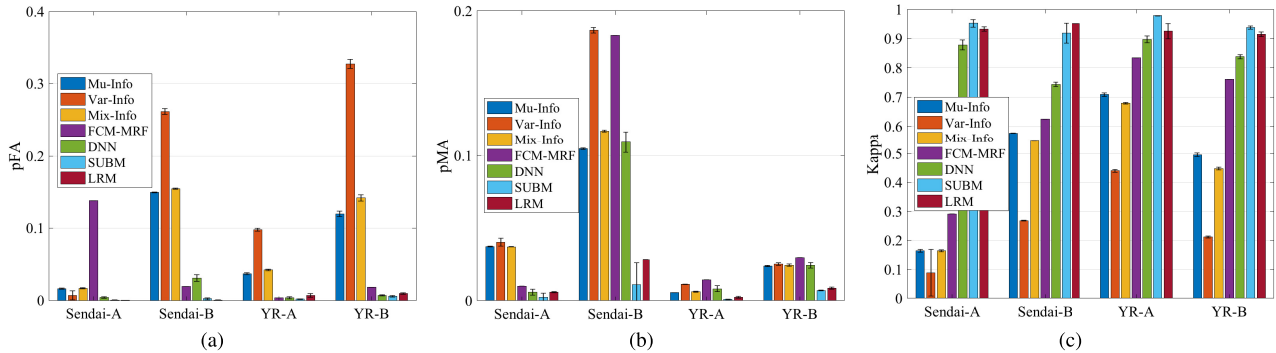

Fig. 3. Quantitative evaluations of the compared methods by (a) pFA, (b) pMA, and (c) Kappa.

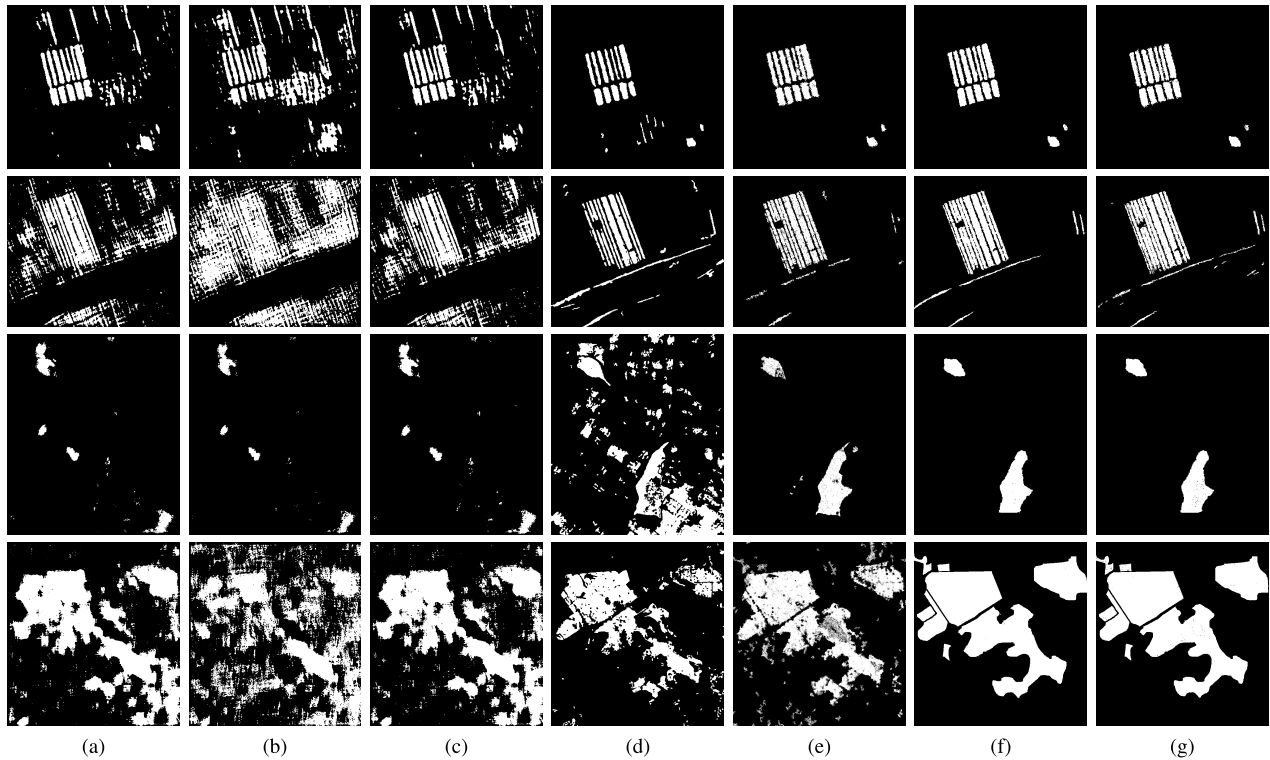

Fig. 4. Visual comparison of (a) Mu-Info, (b) Var-Info, (c) Mix-Info, (d) FCM-MRF, (e) DNN, (f) SUBM, and (g) LRM.

In the following experiments, parameter $C$ is set to 40. The patch size is set as 23, which was determined by experience and trials. In addition, $N$ is set to 2000, i.e., 1000 positive and negative constraint pairs are randomly drawn from the constraint set for metric learning. Both LR-based metric (LRM) and subtraction-based metric (SUBM) with different operators diff$(\cdot, \cdot)$ that are defined in (4) and (5), respectively, are compared in the experiments. To verify the effectiveness of our proposed methods, it is compared with three similarity measurements [2] based on mutual information (Mu-Info), variational information (Var-Info), and mixed information (Mix-Info). In addition, it is also compared with an MRF-based change detection [fuzzy C means (FCM)-MRF] method [7] that introduces the spatial prior into SAR image change detection, and a DNN method [10] that achieves the state-of-the-art performance on SAR image change detection. The performance of the compared methods is evaluated by probabilistic missed alarm (pMA), probabilistic false alarm (pFA), and kappa coefficient, where pFA (pMA) is calculated by the ratios between FA (MA) and the number of unchanged pixels (NC). Since the performance of several compared methods may be interrupted by the random initializations, each compared method runs for ten times, and the performance is evaluated by the means and the standard deviations of the quantitative indexes. To ensure fair comparisons, each compared method was set to an optimal configure. For the M-Info, V-Info, and Mix-Info methods, SVM is employed as the classifier for the final decision. In the training stage, the same number of training samples is used to train the SVM classifier.

TABLE I
RUNNING TIMES OF COMPARED METHODS

| Methods | Mix-Info | FCM-MRF | DNN | SUBM | LRM |
|---|---|---|---|---|---|
| Time (s) | 1004865 | 80 | 395 | 1113982 | 113168 |

The quantitative evaluations on the compared methods are shown in Fig. 3. It is shown in Fig. 3(a) and (b) that FCM-MRF has higher pFA on Sendai-A and higher pMA on Sendai-B, YR-A and YR-B data sets, while other compared methods have higher pMA on all data sets. Instead, our methods achieve lower pFA and pMA than other compared methods. In Fig. 3(c), it is obvious that both our SUBM and LRM achieve better performance than other methods in terms of kappa coefficient. Moreover, between two proposed methods SUBM and LRM, they are comparable on these four data sets. SUBM performs better on Sendai-A, YR-A, and YR-B data sets, while LRM performs better on the Sendai-B data set.

Besides quantitative evaluation, the visual results are also compared, as shown in Fig. 4. It is shown that Mu-Info, Var-Info, and Mix-Info methods are almost invalid for all these four data sets, where they misclassify regions into changed regions on YR-A, YR-B, and Sendai-B data sets and they miss the changed regions on the Sendai-B data set. FCM-MRF obtains most structures of changed regions on YR-A, YR-B, and Sendai-B data sets, but it is invalid on the Sendai-A data set. DNN and our proposed methods can obtain more accurate changed regions than other methods. More specifically, on the YR-B data set (shown in the second row of Fig. 4), compared with DNN, SUBM and LRM can get more complete changed regions, especially the linear-shape changed regions on the bottom of the changed image. In addition, on the Sendai-A and Sendai-B data sets, SUBM and LRM can get more clear structures on changed regions, while DNN has more noisy misclassifications on the changed image.

Finally, we also show the running times of these compared methods. Since Mu-Info, Var-Info, and Mix-Info take similar running times, we only show the running time of the Mix-Info method. We take the Sendai-A data set as an example and the comparisons are given in Table I.

## IV. CONCLUSION

In this letter, a novel spatial metric learning method has been developed for bitemporal SAR images change detection, where spatial context was considered in building the constraint pairs to reduce the effects of speckle and compensate for unavoidable registration errors. Then, the constraint pairs are fed into a metric learning model to train a PSD metric matrix-based max-margin criterion. To verify the effectiveness of the proposed method, we compared the proposed methods with other state of arts on four challenging data sets. The comparison results show that the proposed method outperforms than the other compared methods.